\documentclass[letterpaper]{article}  
\usepackage{aaai17}  %Required
\usepackage{times}  %Required
\usepackage{helvet}  %Required
\usepackage{courier}  %Required
\usepackage{url}  %Required
\usepackage{graphicx}  %Required
\usepackage{color}
\nocopyright{}

\newcommand{\LAAIR}{\textsc{laair}}

\newcommand{\todo}[1]{}
\renewcommand{\todo}[1]{{\color{red} TODO: {#1}}}

\title{\LARGE \bf LAAIR: A Layered Architecture for Autonomous Interactive Robots
}

\author{
Yuqian Jiang\thanks{Equal contribution.}\textsuperscript{1}, Nick Walker\footnotemark[1]\textsuperscript{2}, Minkyu Kim\textsuperscript{3}, Nicolas Brissonneau\textsuperscript{3},\\
\bf{ \Large Daniel S. Brown\textsuperscript{1}, Justin W. Hart\textsuperscript{1}, Scott Niekum\textsuperscript{1}, Luis Sentis\textsuperscript{3}, Peter Stone\textsuperscript{1}} \\
\textsuperscript{1}Department of Computer Science, University of Texas at Austin, Austin, USA\\
\textsuperscript{2}Paul G. Allen School of Computer Science \& Engineering, University of Washington, Seattle, USA\\
\textsuperscript{3}Department of Aerospace Engineering and Engineering Mechanics, University of Texas at Austin, Austin, USA\\
\{jiangyuqian, nickswalker, steveminq, nicolasb\}@utexas.edu\\
\{dsbrown, hart, sniekum\}@cs.utexas.edu, lsentis@austin.utexas.edu, pstone@cs.utexas.edu
}

\begin{document}

\maketitle
\thispagestyle{empty}
\pagestyle{empty}

%%%%%%%%%%%%%%%%%%%%%%%%%%%%%%%%%%%%%%%%%%%%%%%%%%%%%%%%%%%%%%%%%%%%%%%%%%%%%%%%
\begin{abstract}

When developing general purpose robots, the overarching software architecture can greatly affect the ease of accomplishing various tasks. Initial efforts to create unified robot systems in the 1990s led to hybrid architectures, emphasizing a hierarchy in which deliberative plans direct the use of reactive skills. However, since that time there has been significant progress in the low-level skills available to robots, including manipulation and perception, making it newly feasible to accomplish many more tasks in real-world domains. There is thus renewed optimism that robots will be able to perform a wide array of tasks while maintaining responsiveness to human operators. However, the top layer in traditional hybrid architectures, designed to achieve long-term goals, can make it difficult to react quickly to human interactions during goal-driven execution. To mitigate this difficulty, we propose a novel architecture that supports such transitions by adding a top-level reactive module which has flexible access to both reactive skills and a deliberative control module. To validate this architecture, we present a case study of its application on a domestic service robot platform.

\end{abstract}

%%%%%%%%%%%%%%%%%%%%%%%%%%%%%%%%%%%%%%%%%%%%%%%%%%%%%%%%%%%%%%%%%%%%%%%%%%%%%%%%
\section{Introduction}

Researchers have long sought to develop robots that are able to undertake complex tasks autonomously in real-world environments. Early efforts to develop such robots resulted in deliberative systems, in which the robot plans a sequence of actions to achieve a goal, and reactive systems, in which layers of local behaviors react to sensor input \cite{kortenkamp2016}. Hybrid architectures, which layer deliberation and reactivity, emerged as a promising approach to the creation of integrated autonomous robots.

As efforts to improve individual robotic capabilities dominated the research landscape, the outline of hybrid architectures for general purpose robotics has seen few changes in the past two decades. In this time, researchers have made significant progress towards robust robot capabilities such as autonomous localization and navigation \cite{fox1999}, object manipulation \cite{gualtieri2016}, object recognition \cite{krizhevsky2012}. This increased capacity---the fruit of better system modeling, increased computational power, and new tools and techniques---has made it feasible for robots to act intelligently in more dynamic, real-world settings. 

There is a renewed vision that robots will  not only operate autonomously, but do so in challenging environments such as the home or office, where they must regularly interact with humans. Many projects have taken on the grand challenge of creating interactive autonomous robots for use in daily settings \cite{khandelwal17}, and international robotics communities have created competitions such as RoboCup@Home \cite{holz2013} and the World Robot Summit Challenge\footnote{\url{http://worldrobotsummit.org/en/about/}} to encourage research efforts in this direction.

Though some hybrid architectures for general purpose robots were created with human interaction in mind, the extent to which robots operate in populated human environments was neither possible nor expressly accounted for in their design. In a typical layered architecture like 3\textsc{t} \cite{bonasso1997experiences}, skills, which control the connections between sensors and actuators, are placed at the bottom. These low-level control components are invoked and monitored by the middle sequencer layer to achieve planned behaviors. At the top, a task planning layer decomposes the current task into a plan of lower-level behaviors. Given an input task specification, these layers provide the complete loop of planning, monitoring and executing a task. 

This design does not address the desire for robot assistants that constantly interact with people and dynamically receive and execute all kinds of tasks in the environment over extended periods of time. In order to seamlessly respond to human interactions, the top layer must be reactive and maintain direct control of the lower level components, such as a dialogue handling skill and a component that parses commands. 
 
In this paper, we propose a layered architecture for autonomous interactive robots, \LAAIR{}, to facilitate complex tasks in long-term settings and dynamic interactions with humans in real-world domains. The top-level of a \LAAIR{} system is reactive control, which sequences and executes skills in response to the environment and interactions. When the top-level encounters tasks that cannot be statically decomposed, it invokes a deliberative controller which plans and executes actions to accomplish the goal. The bottom level consists of skills which interface with the world. We contrast \LAAIR{} with existing robot architectures and present a case study of applying \LAAIR{} on a mobile robot platform.

\section{Related Work}

The problem of structuring the software of an intelligent robotic system has long been pursued. Early architectures centered on robot planning systems, built around ``sense-think-act" cycles. Shakey, for example, leveraged a STRIPS planner \cite{nilsson1984}. At the time of their development, computational limitations made complex modeling methodologies such as those used for inverse-kinematic motion planning an inaccessibly slow process, and technologies for perception such as object recognition lacked the maturity of modern systems. Partially as a response, reactive systems, which more closely coupled sensing and acting, emerged. In the Subsumption architecture, a well-known example, the robot's behavior is governed by a hierarchy of reactive layers in which control from higher levels subsumes that of lower levels \cite{brooks1986}.

Hybrid architectures draw together deliberative and reactive control, most commonly by placing a high level planner in control of various reactive components. In three layer architectures, this connection is mediated by an executive, commonly a hierarchical state machine, which orchestrates the particular low-level skills used to accomplish a plan \cite{gat1997}. Notable examples of such architectures include \textsc{3t} \cite{bonasso1997experiences}, \textsc{tca} \cite{simmons1994structured} and \textsc{atlantis} \cite{gat1992}.

Our architecture shares many attributes with \textsc{atlantis}. In both architectures, the task planner is only called by the executive control. In \textsc{atlantis}, this design was driven by the need to support asynchronous calls to the slow planning process. Where \textsc{atlantis} makes the planner the primary decision maker for the system, in \LAAIR, the top layer is reactive control. This layer is responsible for decomposing tasks, either by invoking the deliberative control layer or by directly sequencing skills. This is driven by the need to encode both static and dynamic task decomposition and maintain responsiveness during long-term autonomous deployments.

In recent decades, many advancements have been made towards general purpose autonomous interactive robots. Robot Operating System (ROS) has emerged as a dominant software framework in the community, encouraging roboticists to think of their system as an agglomeration of standard, interchangeable components \cite{quigley2009}. Many robotic systems have leveraged ROS and other standardized software components as a foundation on which to improve specific skills or tasks. 

These efforts have laid the groundwork for new approaches to the design of general purpose robot architectures, the challenge \LAAIR{} addresses. While several integrated systems have been designed to address particular challenges for service robots, such as interfaces to natural language commands~\cite{chen2016}, planning in realistic domains~\cite{Tran2017,Hanheide2017}, and real-world scene perception~\cite{beetz2015robosherlock}. Although these systems tackle similar challenges in architecting complex robotic systems, none of these efforts have proposed an overall architecture to support general purpose service robots. \LAAIR{} was designed to organize and reuse interfaces and skills across robots and tasks.

\section{LAAIR}

\begin{figure}
    \centering
    \includegraphics[width=\columnwidth]{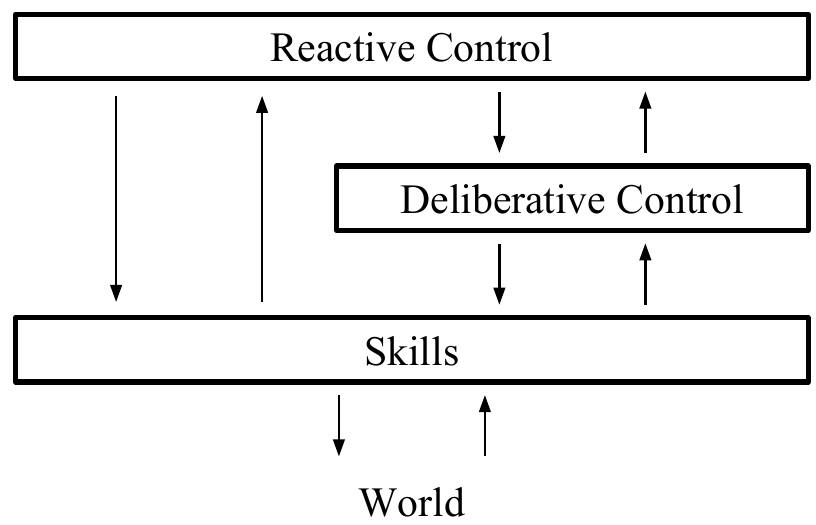}
    \caption{The prototype for a \LAAIR{} system}
    \label{fig:diagram}
\end{figure}

\LAAIR{}, depicted in Figure \ref{fig:diagram}, is a three layer hybrid architecture, consisting of a pool of modular skills, a deliberative control layer that can sequence these skills to achieve goals, and a reactive control layer which drives the system's behavior. The reactive control layer can either directly schedule skills or delegate to the deliberative control layer. Each layer has asynchronous supervisory control over the layers beneath it. This makes it easy to specify systems that remain responsive to human interaction by, for instance, running a user engagement detection skill from the reactive layer, which can then redirect the rest of the system's behavior.

%\commentp{It seems strange to me that the reactive layer doesn't have some task input.  How does the reactive layer know what skills and plans to call?  Is that somehow passed to it indirectly from the world?   From reading the section on skills, it seems that it may be that there is a skill for identifying the current task.  But if so, that's not made explicit.}

\LAAIR{} facilitates code and logic reuse by separating low-level, often robot-specific implementation of skills from robot-independent task structure. The top-level reactive layer simplifies the specification of the top-level scripted behavior of the robot. The deliberative control layer provides the robot with the ability to reason about its environment flexibly as necessary. 

\subsection{Skills}

Skills are the primary interface between the system and the world. They encode robot behaviors, ranging from low-level actions, like moving a joint, to higher level actions, like picking up an object. Everything from perceptual capabilities to the robot's dialog agent are implemented as skills. Skills may accept parameters and are responsible for detecting and reporting their own failure. Importantly, skills must adhere to some uniform interface so they can be directly sequenced by either reactive or deliberative control. This constraint promotes the reuse of skills in different contexts and the portability of the architecture across different platforms. Outside of these constraints, skills may be implemented as best suits their objective---whether it be a procedural program, a state machine or otherwise. 

\subsection{Deliberative Control}

The deliberative control layer is responsible for turning a goal into a sequence of actions that accomplish that goal. These actions should either map directly to skills or be decomposable into a sequence of skills. In addition to the actual process of generating the sequence of skills, the deliberative control layer is responsible for monitoring the execution of the sequence, intervening when it determines that resequencing or other corrective action is necessary. The deliberative control layer reports major milestones in execution, such as completed actions or exceptional behavior to the reactive control layer, so that they may be optionally handled in a task-specific manner.

Because \LAAIR{} gives the reactive control layer discretion on the specification of goals, the architecture can be instantiated with a broad range of deliberative components, or even use different components within the course of accomplishing a single task. Further, goal specification flexibility allows the reactive control layer to statically decompose a complicated task into several goals to limit the computational expense of deliberation.

\subsection{Reactive Control}

The reactive control layer is the primary executive in the system. It contains a high-level representation of the robot's task, for instance, a hierarchical finite state machine. In simple cases, this layer uses a static representation of the task to sequence skills and handle contingencies for different execution outcomes. In complicated tasks where static decomposition is infeasible, this layer produces a goal or goals that can be dynamically resolved by the deliberative control layer into a sequence of skills. The reactive control layer is responsible for monitoring and handling the outcomes of the skills that it directly calls, as well as for supervising deliberative execution. This gives the layer overarching control of the system, allowing it to preempt execution when, for instance, a human engages with the robot.

The reactive control layer also facilitates the hand-specification of actions for static parts of a task, while simultaneously supporting the use of deliberative control for instances where the task is dynamic. Because executive control over the rest of the system is maintained from this layer, the robot's behavior is always attributable to some portion of its top level representation. This makes \LAAIR{} systems easier to understand at runtime and easier to debug during development.

\begin{figure*}
    \centering
    \includegraphics[width=\textwidth]{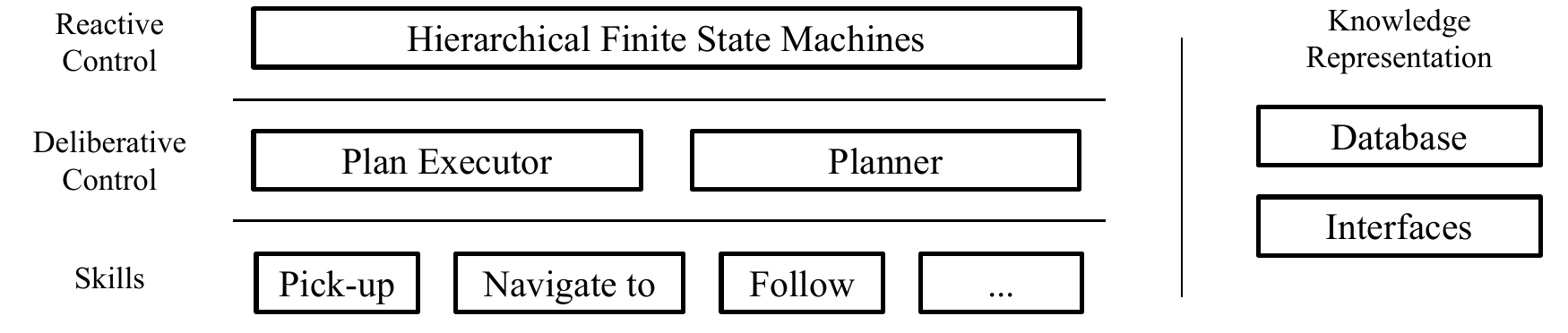}
    \caption{Functional components of our \LAAIR{} implementation for HSR.}
    \label{fig:hsr_system_diagram}
\end{figure*}

\section{Case Study}

We show the instantiation of \LAAIR{} on a Toyota Human Support Robot (HSR) as part of an entry into the 2018 RoboCup@Home Domestic Standard Platform League. The system is designed to execute both highly prescribed tasks, like guiding a person from a vehicle to an apartment, as well as open-ended tasks, like servicing requests that could require the robot to execute any number of actions in arbitrary order.

HSR is a domestic robot platform equipped with an omnidirectional base, an arm, stereo and RGB-D vision systems, a speaker as well as a microphone array. The software stack of HSR is based on ROS. In this instantiation of the architecture, depicted in Figure~\ref{fig:hsr_system_diagram}, the reactive control layer is provided by hierarchical finite state machines, deliberative control leverages a symbolic task planner, and skills range from custom implementations to open source components.

\subsection{Reactive Control}

We use SMACH\footnote{\url{http://wiki.ros.org/smach}} to implement hierarchical finite state machines describing each RoboCup@Home task. Many of the tasks follow a static series of steps, so they are implemented purely by sequencing skills. When the robot is required to accept natural language commands and dynamically decompose the task, we enter a sub-state machine which calls skills to formulate a goal that can be delegated to the deliberative control layer. If the plan execution encounters errors, a sub-state machine in the reactive control layer engages to either specify another goal or execute a remedial sequence of skills. The reactive layer connects a command dialogue skill to a constantly-running wrist tap detection skill to enable the robot to accept new tasking.

%\commentp{It's still not clear where the central control loop resides.  How does the reactive layer accept new tasks?}

\subsection{Deliberative Control}

We implement the deliberative control of our robot using a task planner and a plan executor. The task planner is based on Answer Set Programming (ASP) \cite{lifschitz2002answer}, and we use the answer set solver \textsc{clingo} to generate plans \cite{gebser2011potassco}. The plan executor is responsible for invoking the planner after receiving a goal, calling the corresponding skills, and monitoring the execution. 

%\commentn{would be nice to add: What does it report up, when does it error out?}

\subsection{Knowledge Base}
Besides the control layers, central to our system is a knowledge base module that stores a concept network and situated knowledge about the domain. We implement the knowledge base with a relational database managed by MySQL. The knowledge base represents entities and attributes. Each entity is assigned an ID, and can represent an abstract concept or a concrete object in the environment. Attributes describe properties and relations of entities. 

The knowledge base is accessed by all layers of control in the architecture: the reactive executive writes and reads the identity of the operator and other task-level information; the deliberative control plans over domain knowledge; the skill components retrieve relevant information and update state as necessary.

\subsection{Skills}

 We leverage the action server/client abstraction available in ROS to implement skills as processes which can be provided a goal, send feedback to a supervisor, accept cancellation requests and return outcomes. In exceptional cases, such as skills that return significant amounts of data, we implement skills as library functions to avoid interprocess communication overhead. Adopting this standardized interface for most skills simplifies how the deliberative or reactive layers interacts with them. 
 
 The selection of skills described in this section highlight the wide range of implementation in this layer.

\subsubsection{Command Dialog agent}

This skill manages a dialogue with the operator, parses the command, and resolves the task type and parameters. The commands are generated from the fixed grammar used in the Robocup@Home Competition\footnote{\url{https://github.com/kyordhel/GPSRCmdGen}}. Our current system understands 32 high-level domestic tasks, such as finding people, answering questions, and delivering objects, which can be sequenced in any order to form a wide variety of complex commands.

Our implementation of the skill transcribes an utterance using the Google Cloud Speech-to-Text API. We then build a parse tree by expanding the production rules to all possible sentences. For each command, we traverse the parse tree to match the task type, such as ``navigate to'', and to fill in task parameters, such as ``dining table''.

After fully parsing a command, we resolve coreferences by searching backwards in the sentence to find the closest name or object. If a coreference cannot be resolved, or a command is incomplete, we engage in a correction dialogue to attempt to recover the missing information. Based on the semantics and type of missing information, the robot selects from a library of templates in order to ask an appropriate clarification question and resolve the parse.

%\commentp{Is this skill always running by default?  Or whenever some other skill isn't?  Can this be seen as the "main" process?}

%\input{dialog_agent_long.tex}

%\subsubsection{Semantic Navigation}

%\todo{More}

%\subsubsection{Perception}

%\subsubsection{Object Manipulation}

\subsubsection{Detection of movable objects }
We assume that objects which the robot may need to move---to clear a path for navigation, for instance---include a curved surface and at most two accumulated degrees of freedom. We detect potentially movable objects through two independent methods:

The first runs continuously, recording a 2D ground map of the environment based on laser and depth readings and comparing it with incoming readings. This method is robust to small changes to the environment, making it appropriate for detecting whether small items such as books, bags or cans on the floor have moved. The second searches for cylindrical surfaces in the scene and classifies them as part of potentially movable objects. This method is more efficient for identifying bigger obstacles such as rolling chairs or doors as it relies on their degrees of freedom.

%\begin{figure}[h]
%    \centering
%        {\includegraphics[width=0.8\linewidth]{structure.png}}
%    \caption{Detection strategies of moveable objects}
%       \label{Figure:Manipulation_approach}
%\end{figure}

\subsubsection{Person Following}

\begin{figure}[t]
    \centering
        {\includegraphics[width=0.8\linewidth]{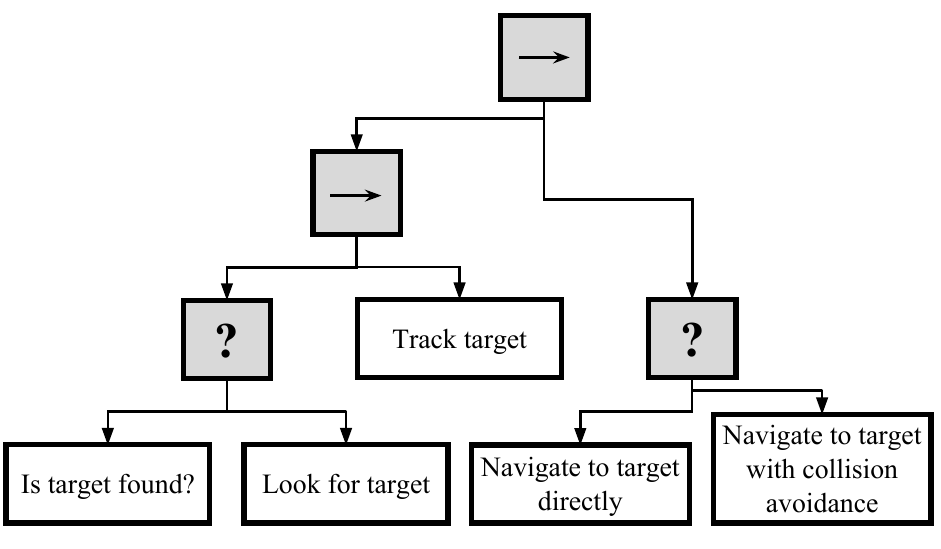}}
    \caption{Behavior tree for person following skill. Arrows indicate sequence nodes and question marks indicate fallback nodes.}
       \label{Figure:BT_framework}
\end{figure}

We divide the person following skill into three steps; detecting the target, tracking the target with the robot's head, and navigating towards the target for following. These behaviors must be coordinated based on the state of the robot and its surroundings. We model this task as a behavior tree, which provides a compact representation of how the robot should move between actions while executing the behavior. Our implementation leverages a ROS behavior tree framework, described in \cite{colledanchise2017behavior}. Our behavior tree is designed with two fallback nodes and two sequence nodes as shown in Figure \ref{Figure:BT_framework}. 

%\subsection{Demonstration}

%\input{hsr_demonstration.tex}

%\subsection{BWIBot}

%\subsubsection{Skills}

%\subsubsection{Planning Domain}

%\subsubsection{Demonstration}

%The robot is delivering a message when it is interrupted and asked for directions. The robot gives directions to the operator, then assumes its previous goal.

\section{Future Work}

Our case study has demonstrated the potential for \LAAIR{} to address the needs of a general purpose robot operating in a human-populated environment. We are currently implementing \LAAIR{} on an office robot platform to demonstrate how the architecture enables a high degree of software portability across service robots. In further instantiations, we plan to demonstrate the architecture's ability to handle interuptions and concurrency.

\section*{Acknowledgements}

This work has taken place in the Learning Agents Research
Group (LARG) at UT Austin.  LARG research is supported in part by NSF(IIS-1637736, IIS-1651089, IIS-1724157), Intel, Raytheon, and Lockheed Martin.  Peter Stone serves on the Board of Directors of Cogitai, Inc. The terms of this arrangement have been reviewed and approved by the University of Texas at Austin in accordance with its policy on objectivity in research.

\bibliographystyle{aaai}
\bibliography{references}

\end{document}